\documentclass{IEEEoj}
\usepackage{cite}
\usepackage{amsmath,amssymb,amsfonts}
\usepackage{algorithmic}
\usepackage{graphicx,color}
\usepackage{textcomp}
\usepackage{graphicx}
\usepackage{subfigure}
\usepackage{wrapfig}
\usepackage{tabularx}
\usepackage{multirow}
\usepackage{color}
\usepackage[table]{xcolor}
\usepackage{colortbl}
\usepackage{tikz}
\usepackage{xcolor}
\usepackage{adjustbox}
\usepackage[figuresleft]{rotating}
\usepackage{caption}

\usepackage[T1]{fontenc}

\newcommand{\key}{\rotatebox{90}}

\ifCLASSINFOpdf
\else
\fi
\usepackage{amsmath}
\usepackage[cmintegrals]{newtxmath}
\makeatletter
\AtBeginDocument{\definecolor{ojcolor}{cmyk}{0,0,0,1}}
\makeatother

\begin{document}

\title{A new Hybrid Model of Generative Adversarial Network and You Only Look Once Algorithm for Automatic License-Plate Recognition}



\author{
	Behnoud~Shafiezadeh\IEEEauthorrefmark{1}, 
	Amir~Mashmool\IEEEauthorrefmark{2},  
	Farshad~Eshghi\IEEEauthorrefmark{3}, 
	Manoochehr~Kelarestaghi\IEEEauthorrefmark{1}
}

\affil{
	\IEEEauthorrefmark{1}Electrical and Computer Engineering, Kharazmi University, Tehran, 100190, Iran \\
	\IEEEauthorrefmark{2}Faculty of Mathematics and Computer Science, University of Bremen, Bremen, 28359,Germany \\
	\IEEEauthorrefmark{3}School of Computing, University of the Fraser Valley, Abbotsford, V2S 7M8, Canada
}



\begin{abstract}
Automatic License-Plate Recognition (ALPR) plays a pivotal role in Intelligent Transportation Systems (ITS) as a fundamental element of Smart Cities. However, due to its high variability, ALPR faces challenging issues more efficiently addressed by deep learning techniques. In this paper, a selective Generative Adversarial Network (GAN) is proposed for deblurring in the preprocessing step, coupled with the state-of-the-art You-Only-Look-Once (YOLO)v5 object detection architectures for License-Plate Detection (LPD), and the integrated Character Segmentation (CS) and Character Recognition (CR) steps. The selective preprocessing bypasses unnecessary and sometimes counter-productive input manipulations, while YOLOv5 LPD/CS+CR delivers high accuracy and low computing cost. As a result, YOLOv5 achieves a detection time of 0.026 seconds for both LP and CR detection stages, facilitating real-time applications with exceptionally rapid responsiveness. Moreover, the proposed model achieves accuracy rates of 95\% and 97\% in the LPD and CR detection phases, respectively. Furthermore, the inclusion of the Deblur-GAN pre-processor significantly improves detection accuracy by nearly 40\%, especially when encountering blurred License Plates (LPs).To train and test the learning components, we generated and publicly released our blur and ALPR datasets (using Iranian license plates as a use-case), which are more representative of close-to-real-life ad-hoc situations. The findings demonstrate that employing the state-of-the-art YOLO model results in excellent overall precision and detection time, making it well-suited for portable applications. Additionally, integrating the Deblur-GAN model as a preliminary processing step enhances the overall effectiveness of our comprehensive model, particularly when confronted with blurred scenes captured by the camera as input.
\end{abstract}



\maketitle

\section{INTRODUCTION}
\label{introduction}

Rising population trends are fueling a surge in vehicular usage. Consequently, governments are increasingly pressed to implement sophisticated controls within transportation systems to enhance infrastructure quality. Advancements in computer science, particularly in machine vision, are making significant strides and finding practical applications in Intelligent Transportation Systems (ITS). Machine vision technology harnesses the capabilities of cameras to extract valuable data, including information about license plates, vehicles, and even individuals. Notably, each vehicle possesses a distinct license plate (LP) that functions as its unique identifier. The process of identifying these plates falls under one of ITS's key applications: Automatic License Plate Recognition (ALPR).
Vehicle detection through LP identification has applications in traffic management (e.g., open-roads toll payment, restricted traffic zones, parking systems, and vehicle access control), traffic violations, and criminal investigations involving vehicles.

\begin{figure*}[h!]
	\centering
	\captionsetup{justification=centering}
	\includegraphics[width=\textwidth]{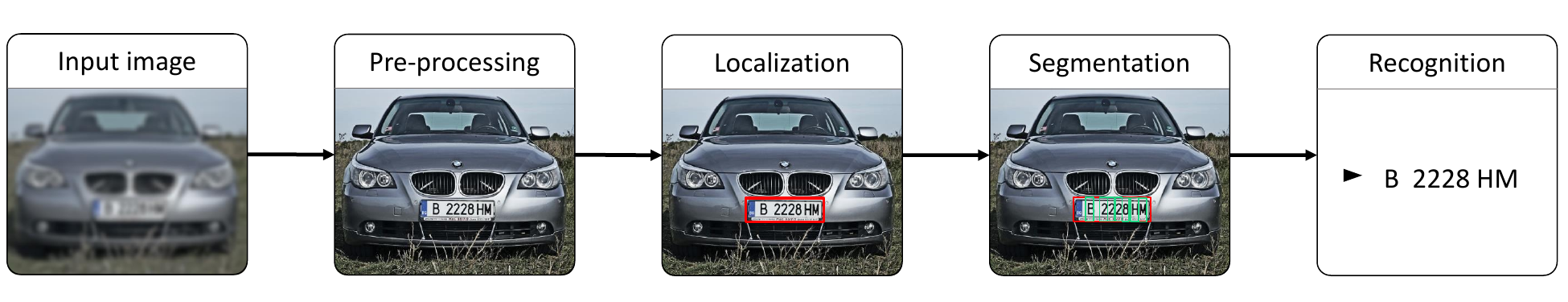}
	\caption{ALPR system steps.}
	\label{fig:ALPR_system}
\end{figure*}

Real-time capability stands as a paramount factor in transportation systems, particularly when faced with escalating vehicle volumes within traffic control systems. The imperative for instantaneous decision-making for each car intensifies. ALPR systems, powered by cutting-edge machine vision algorithms, facilitate rapid and precise decisions, ensuring swift actions are taken within the shortest time frame possible.
As typical LPs carry graphic-only information, ALPR is conventionally an image-processing task. In general, ALPR systems comprise four steps of preprocessing, License-Plate Detection (LPD), Character Segmentation (CS), and Character Recognition (CR). As illustrated in Fig.~\ref{fig:ALPR_system} the quality of the image is improved in the preprocessing step. The LPD step localizes the LP in the already improved image. The segmented characters of the LP image in the CS are then recognized in the CR step. The steps mentioned above might be eliminated and/or implemented in a smaller number of separate steps. For instance, in some ALPR systems, the CS and CR tasks can be combined into one step, while the preprocessing step might not exist.

\par There are two main categories of ALPR algorithms in the literature: Conventional and Learning-based. With a long research history, the former category includes complex mathematical techniques that explore the specific graphical (e.g., geometrical or statistical) features of LPs in an image for recognition purposes. In mathematical approaches, a task can usually be viewed and implemented as a set of sub-tasks to provide insight and tractability. However, Conventional methods have high computational complexities. Examples of mathematical image object detection and classification techniques are histogram, edge detection, and morphological image processing ~\cite{P7_2016}. On the other hand, learning-based ALPR algorithms perform the same task by extracting high-level features using some form of iterative parameters adjustment of simple computational units configured as in layers (e.g., Multi-Layer Perceptron (MLP) ~\cite{P5_2011}). ~\cite{P11_2020}Compared to their conventional counterparts, the procedures are more end-to-end, and the details are less transparent. Learning-based ALPRs need a large volume of training data and are associated with high-offline (i.e., at the training phase) and relatively low-online (i.e., at the execution phase) time consumption. In recent years, learning-based ALPRs are becoming more and more attractive from practicality and performance viewpoints due to their high parallel programming implementation potential and ever-increasing availability of computationally powerful GPUs.

\par As any other imaging classification method, the ALPR performance is impacted by the input image's varying quality and appropriateness. However, ALPR systems face extra challenges as their input images are usually shots generated from captured videos of mobile objects in uncontrolled-open spaces with the following peculiarities: 
\begin{itemize}
	\item different resolution levels in capture or storage 
	\item harsh weather conditions
	\item uneven (flash effect) illuminations 
	\item LP object sizes 
	\item angled-view/multiple/partially-occluded LP objects
	\item different locations of the LP objects in the image
	\item  Blurriness due to movement.	
	\item Specifics of LPs e.g., size, color, and characters, in each country
\end{itemize}

\subsection{Motivation and Contribution} 
\label{Motivationpart}
Most of the ALPR systems introduced in the literature seek efficient applicability in the transportation industry. Thus, they generally employ datasets generated by fixed-setup imaging systems with known specifications and, to some extent controlled subject appearance (for instance, highway and parking ALPRs), resulting in not very high variability in image circumstances and parameters. While these ALPR systems might be sufficient for transportation industry applications, extending ALPR applicability to other fields challenges their merits. In particular, for instance, in security applications, the input images might have come from multiple imaging setups/configurations taken in an ad hoc manner. As a result, more often than not, many earlier-mentioned image challenges show up together. Furthermore, reliability is critical in these applications, and depending on the application at hand, the recognition test response time might also become a concern. 

\par Targeting a broader ALPR applicability,   a ALPR system capable of being adequately trained with images of more versatile quality and circumstance is proposed, delivering a low error rate and good test response time. The highlights of the proposition are:

\begin{itemize}
	\item \textbf{Employment of the state-of-the-art YOLOv5 structures} for its proven power for detecting small-size objects such as LPs and characters, and being computationally lightweight, making it suitable for embedded and near-real-time applications.
	\item \textbf{Generation of two comprehensive and representative datasets} one to compensate for the lack of a public dataset that accommodates Iranian LPs in different circumstances of angle, zoom, resolution, and subject multiplicity, and various environmental conditions of weather, illumination, light, and noise, and the other one to specifically train a deblurrer structure. 	
	\item \textbf{Proposing a selective GAN-deblurring in the preprocessing step} to mitigate, if tested to be necessary, the effect of blurring, as it is a typical performance degrading phenomenon in shooting moving subjects or shaky shootings.
	
\end{itemize}

\par The rest of the paper is organized as follows. Section~\ref{Review} discusses the state-of-the-art related works categorized into conventional, hybrid, and learning-based methods. Also, a self-explanatory, quick reference is provided in this section. The proposed ALPR system is explained in section~\ref{Methodolgy}, following a brief background regarding GAN-based debluring and YOLO-based object detection. The details of the generated ALPR and Deblur datasets are provided in section~\ref{Dataset}. The simulation setup and performance evaluation results appear in section~\ref{Simulation} followed by concluding remarks in section~\ref{Conclusion}.

\section{Related Works}
\label{Review}

ALPR related works are grouped into three categories, conventional, hybrid (i.e., combined conventional and learning), and overall learning-based methods.

\subsection{Conventional approaches}
\label{Conventional}

\par Starting with conventional ALPRs, in \cite{P1_2009}, at the preprocessing step, to increase the contrast of LP regions and to avoid false-positive LPs, two different image enhancement methods based on intensity variance and edge density are used. Then, a matched filter that models the vertical edge density and a Multimodal Neighborhood Signature (MNS) method that identifies an LP's color information are employed to detect the LP candidate regions. There is no provision for CR in this paper.
Similar to the previous work, \cite{P2_2012} only performs LPD by applying a Sobel edge filter in the vertical direction on the gray level input image to detect the LP edges. As the work concerns Iranian LPs, after the binarization, the blue region at the beginning of the LP will be searched for.
To smooth background areas, especially in LP regions, \cite{P3_2013} uses region-based filtering in the preprocessing step. In the second step, a Sobel and morphological filter is employed for extracting vertical edges and candidate regions. While using a morphological filter increases the accuracy, the region-based filtering reduces the time response.
Rashedi et al. \cite{P4_2018}, a four-method hierarchical LPD algorithm comprising cascade classifier, edge-based, color-based, and contrast-based methods were proposed. First, the illumination level of the input image is compared against pre-determined thresholds, and based on the result, the parameters and the application order of the methods in the hierarchy are specified. Indeed, the algorithm tries to synergize the advantages of above mentioned four localization methods.

\subsection{Hybrid Methods}
\label{Hybrid}
\par The availability of a large volume of vehicle image data and the maturity of learning-based image processing techniques made it possible to implement the ALPR methods in hybrid. Assuming that the input image is a single LP, \cite{P5_2011} proposes an algorithm that spans the CS and CR steps of ALPR. First, Gaussian and Laplacian filters decrease the effect of noise and illumination non-uniformity in the LP's region. Then, the characters are segmented using geometrical operations, their features are extracted and decreased in number using the PCA algorithm and finally recognized by two feed-forward NNs.
Authors in \cite{P6_2014} combine modified template-matching and pixel colors to identify the color strips in Iranian LPs and use the LP's shape and aspect ratio to specify the LP's candidate region. Afterward, The projection-based algorithm is applied to the LP candidate regions to segment the characters of LP. The characters are then recognized using a hybrid Decision Tree-SVM classifier.
\cite{P7_2016} applies a Prewitt edge detection filter on the grey-level version of the input image to localize the LP. After that, by morphological operations, the characters of the LP are first segmented and then recognized by a hybrid KNN-SVM classifier.
In the LP step, \cite{P8_2017} applies adaptive-threshold filtering on the gray-level input image to specify the region of LP candidates. Then, the LP regions are detected by using Connected Component Analysis (CCA) and Random Sampling Consensus (RANSAC). Moreover, an SVM classifier determines the quality of the LP (dirty, medium, and clean). Finally, The characters of the LP are segmented and recognized by applying adaptive-threshold filtering prior to a CCA features extraction followed by an SVM classifier.
As the input image in work done by \cite{P9_2018} is an LP image, there is no LP detection step therein. Instead, The input LP image is first enhanced at the pre-processing step by normalizing the plate size and applying a wiener filter. Then, an NN detects poorly illuminated LPs to be histogram equalized. Finally, contrast enhancement is performed using a particular transform function. Next, local binarization distinguishes black characters from white backgrounds, and the vertical and horizontal projections segment the LP characters. Finally, In the last step, an MLP classifier recognizes the characters.

\subsection{Learning-based methods}
\label{learning-based}
Since a decade ago, fully-learning-based methods have made their way into the ALPR field. In light of the advancements in the YOLO algorithm, as detailed in section~\ref{obj}, this algorithm and its various versions have found application in the domain of ALPR.
\cite{P37_lastpaper_eccv2018} proposes a convolution-based architecture in which recognition and detection are implemented in one step (with sharing features). The approach introduces a significant detection speed improvement compared to the other methods with separate LP and character detection. The latter uses the publicly available CCPD dataset containing car images captured manually at a city parking in China.
\cite{P10_2019} applies a sliding window to the input image and then employs 36 simplified YOLO structures (tiny-YOLO) with a reduced number of convolutional layers, each adapted to detect a distinct character or LP. 
The work in \cite{P11_2020} is explicitly proposed for the CR step. The input image is assumed to be a single LP image. The high-level features of the input LP image are extracted using an Auto-Encoder neural network. The extracted features set is then applied to 8 Convolutional Neural Networks (CNNs) concurrently, each responsible for detecting one of 8 digits of the input LP.
\cite{P12_2020} proposes a YOLOv3 structure for detecting LPs in an input image. The same authors complement their work in \cite{P12_2020} by adding another YOLOv3 structure for CR \cite{P13_2020}.
The ALPR in \cite{P14_2020} detects LPs in two passes using a Fast YOLO network. The front/rear view of cars in the input images is cropped on the first pass, and on the second pass, the LPs are detected. The detected LPs are fed to a second Fast YOLO for CR. The work is done on LPs from three different countries.
As the ALPR method introduced in \cite{P14_2020} targets multinational LPs, the layouts of LPs detected by a tiny-YOLOv3 network are determined using a layout detection network given the LP's number of lines of characters. A second YOLOv3-SPP network is then employed for CR purposes in \cite{P15_2020}. While the layout detection is claimed to have been done on LPs from 17 countries, the ALPR accuracy results are just reported for the Korean LPs.  
Aiming at real-time performance in embedded systems (with limited memory and clock rates), the proposed ALPR in \cite{P16_2020} uses a tiny-YOLOv3 for LPD and a 36-class CNN structure for CR without any CS step.

A summary of the related work is presented as an ALPR quick reference in Table~\ref{table:literature}.

\begin{sidewaystable*}
	\renewcommand{\arraystretch}{0.9}	
	\caption {ALPR Quick Reference} 
	\label{table:literature}
	\centering
	
	\resizebox{0.87 \textwidth}{!}{
		\begin{tabular}{|c|c|c|c|c|c|c|c|}
			\cline{2-8}
			\multicolumn{1}{c|}{\cellcolor{black}} & 
			\multirow{2}{*}{\tiny \bf Ref} & 
			\multicolumn{4}{ c|}{\tiny \bf Step} &
			
			\multirow{2}{*}{\tiny \bf Dataset / Country} &
			{\tiny \bf Performance} 
			
			\\ \cline{3-6} 
			\multicolumn{1}{c|}{\cellcolor{black}} & 
			&
			
			{\tiny \bf Pre-process} & {\tiny \bf Localization} & {\tiny \bf Segmentation} & {\tiny \bf Recognition} & & {\tiny \bf (LPD / CS / CR / Overall)}
			
			\\ \hline  &

			\multirow{2}{*}{\tiny \cite{P1_2009}} & \multirow{2}{*}{\tiny vertical edge density} & {\tiny edge (match filter)}& \multirow{2}{*}{\tiny NA} & \multirow{2}{*}{\tiny NA} & \multirow{2}{*}{\tiny NP / Iran \textsuperscript{\tiny\fontsize{3pt}{6pt}\selectfont{1}}} & \multirow{2}{*}{\tiny NA ~/~ NA ~/~ NA ~/~ NA }  \\ & & & {\tiny + MNS (color-based model) } & & & & 
			
			\\ \cline{2-8}  \multirow{4}{*}{
				\begin{tikzpicture}[remember picture,overlay]
					\node[fill,rectangle,top color={white!50!red},bottom color={white!50!red},minimum width=0.64cm,minimum height=3.2cm] (test){\key{\tiny \bf Conventional}};
				\end{tikzpicture}
				
			}  &
			
			\multirow{2}{*}{\tiny \cite{P2_2012}} & \multirow{2}{*}{\tiny \tiny NA} & \multirow{2}{*}{\tiny edge (Sobel filter)} & \multirow{2}{*}{\tiny NA} & \multirow{2}{*}{\tiny NA} & \multirow{2}{*}{\tiny NP / Iran \textsuperscript{\tiny\fontsize{3pt}{6pt}\selectfont{1}}} & {\tiny day: 94.0\% /~ NA ~/~ NA ~/~ NA} \\ & & & & & & & {\tiny night: 96.6\% /~ NA ~/~ NA ~/~ NA} 
			
			\\ \cline{2-8} \multirow{13}{*}{
				\begin{tikzpicture}[remember picture,overlay]
					\node[fill,rectangle,top color={white!50!red},bottom color={white},minimum width=0.64cm,minimum height=3.2cm] (test){};
				\end{tikzpicture}
				
			} &
			
			\multirow{2}{*}{\tiny \cite{P3_2013}}  & {\tiny region-based filtering } & {\tiny edge + morphological + } & \multirow{2}{*}{\tiny NA} & \multirow{2}{*}{\tiny NA} & \multirow{2}{*}{\tiny NP / Iran \textsuperscript{\tiny\fontsize{3pt}{6pt}\selectfont{1}}} & \multirow{2}{*}{\tiny 92.0\% /~ NA ~/~ NA ~/~ NA} \\ & & {\tiny edge density} & {\tiny geometrical} & & & &
			
			\\ \cline{2-8} &
			
			\multirow{2}{*}{\tiny \cite{P4_2018}} & \multirow{2}{*}{\tiny NA } & {\tiny color + edge + contrast +} & \multirow{2}{*}{\tiny NA} & \multirow{2}{*}{\tiny NA} & \multirow{2}{*}{\tiny NP / Iran \textsuperscript{\tiny\fontsize{3pt}{6pt}\selectfont{1}}} & \multirow{2}{*}{\tiny 98.4\% /~ NA ~/~ NA ~/~ NA} \\ & & & {\tiny cascade classifier + illumination threshold based } & & & &
			
			\\ \cline{2-8}  &
			
			\multirow{2}{*}{\tiny \cite{P5_2011}} & {\tiny Gaussian low-pass filter +} & \multirow{2}{*}{\tiny NA} & \multirow{2}{*}{\tiny geomtrical manipullation } & \multirow{2}{*}{\tiny PCA + NN} & \multirow{2}{*}{\tiny NP / Iran \textsuperscript{\tiny\fontsize{3pt}{6pt}\selectfont{1}}} & \multirow{2}{*}{\tiny NA ~/ 94.0\% / 90.5\% /~ NA}  \\  & & {\tiny Laplacian transformation} & & & & &
			
			\\ \cline{2-8} &
			
			\multirow{2}{*}{\tiny \cite{P6_2014}} & \multirow{2}{*}{\tiny NA} & \multirow{2}{*}{\tiny color-based + Template matching} & \multirow{2}{*}{\tiny edge + morphological + projection} & \multirow{2}{*}{\tiny decision tree + SVM} & \multirow{2}{*}{\tiny NP / Iran \textsuperscript{\tiny\fontsize{3pt}{6pt}\selectfont{1}}} & \multirow{2}{*}{\tiny 96.6\% / 95.8\% /~ NA ~/ 92.6\%} \\ & & & & & & &
			
			\\ \cline{2-8} &
			\multirow{2}{*}{\tiny \cite{P7_2016}}  & \multirow{2}{*}{\tiny NA} & \multirow{2}{*}{\tiny Prewitt edge detection } & \multirow{2}{*}{\tiny morphology } & \multirow{2}{*}{\tiny KNN + SVM} & \multirow{2}{*}{\tiny NP / Iran \textsuperscript{\tiny\fontsize{3pt}{6pt}\selectfont{1}}} & \multirow{2}{*}{\tiny 96.0\% / 95.2\% / 7.0\% /~ NA} \\ & & & & & & & 
			
			\\ \cline{2-8} &
			\multirow{2}{*}{\tiny \cite{P8_2017}} & \multirow{2}{*}{\tiny NA} & \multirow{2}{*}{\tiny adaptive threshold + CCA + SVM} & \multirow{2}{*}{\tiny adaptive threshold + CCA} & \multirow{2}{*}{\tiny SVM} & \multirow{2}{*}{\tiny NP / Iran \textsuperscript{\tiny\fontsize{3pt}{6pt}\selectfont{1}}} & \multirow{2}{*}{\tiny 98.7\% / 99.2\% / 97.6\% /~ NA} \\ & & & & & & &
			
			\\ \cline{2-8} &
			
			\multirow{2}{*}{\tiny \cite{P9_2018}} & {\tiny low-pass Weiner + NN +} & \multirow{2}{*}{\tiny NA} & {\tiny binarization +} & \multirow{2}{*}{\tiny MLP} & \multirow{2}{*}{\tiny NP / Iran \textsuperscript{\tiny\fontsize{3pt}{6pt}\selectfont{1}}} & \multirow{2}{*}{\tiny NA ~/~ NA ~/ 91.2\% /~ NA}  \\ & & {\tiny histogram equalization} & & {\tiny horizontal and vertical projections} & & &
			
			\\ \cline{2-8} &
			
			\multirow{2}{*}{\tiny \cite{P37_lastpaper_eccv2018}} & \multirow{2}{*}{\tiny NA} &  \multicolumn{3}{c|}{\multirow{2}{*}{\tiny RPnet}} & \multirow{2}{*}{\tiny CCPD / China \textsuperscript{\tiny\fontsize{3pt}{6pt}\selectfont{6}}} & \multirow{2}{*}{\tiny 94.5\% /~ NA ~/ 95.5\% /~ NA}  \\ & & & \multicolumn{3}{c|}{} & & 
			
			\\ \cline{2-8} &
			
			\multirow{2}{*}{\tiny \cite{P10_2019}} & \multirow{2}{*}{\tiny NA} &  \multicolumn{3}{c|}{\multirow{2}{*}{\tiny tiny-YOLO + sliding window}} & \multirow{2}{*}{\tiny AOLP / Taiwan \textsuperscript{\tiny\fontsize{3pt}{6pt}\selectfont{2}}} & \multirow{2}{*}{\tiny 98.2\% /~ NA ~/ 78.0\% /~ NA}  \\ & & & \multicolumn{3}{c|}{} & &

			\\ \cline{2-8} &
			\multirow{2}{*}{\tiny \cite{P11_2020}} & \multirow{2}{*}{\tiny NA} & \multirow{2}{*}{\tiny NA} & \multirow{2}{*}{\tiny NA} & \multirow{2}{*}{\tiny AutoEncoder + CNN} & \multirow{2}{*}{\tiny NP / Iran \textsuperscript{\tiny\fontsize{3pt}{6pt}\selectfont{1}}} & \multirow{2}{*}{\tiny NA ~/~ NA ~/ 96.0\% /~ NA}  \\ & & & & &  & &			
			
			\\ \cline{2-8} &
			\multirow{2}{*}{\tiny \cite{P12_2020}} & \multirow{2}{*}{\tiny NA} & \multirow{2}{*}{\tiny YOLOv3} & \multirow{2}{*}{\tiny NA} & \multirow{2}{*}{\tiny NA} & \multirow{2}{*}{\tiny NP / Iran \textsuperscript{\tiny\fontsize{3pt}{6pt}\selectfont{1}}} & \multirow{2}{*}{\tiny 98.0\% /~ NA ~/~ NA ~/~ NA} \\ & & & & &  & &
			
			\\ \cline{2-8} &
			\multirow{2}{*}{\tiny \cite{P13_2020}} & \multirow{2}{*}{\tiny NA} & \multirow{2}{*}{\tiny YOLOv3} & \multicolumn{2}{c|}{\multirow{2}{*}{\tiny YOLOv3}} & \multirow{2}{*}{\tiny NP / Iran \textsuperscript{\tiny\fontsize{3pt}{6pt}\selectfont{1}}} & \multirow{2}{*}{\tiny NA ~/~ NA ~/~ NA ~/ 95.5\%}  \\  & & & &  \multicolumn{2}{c|}{} & & 
			
			\\ \cline{2-8}  &
			\multirow{3}{*}{\tiny \cite{P14_2020}} & \multirow{3}{*}{\tiny NA} & \multirow{3}{*}{\tiny Fast YOLO} & \multicolumn{2}{c|}{\multirow{3}{*}{\tiny Fast YOLO }} & {\tiny SSIG / Brazil \textsuperscript{\tiny\fontsize{3pt}{6pt}\selectfont{3}}} & {\tiny NA ~/~ NA ~/~ NA ~/ 92.4\% }   \\  & & & & \multicolumn{2}{c|}{} & {\tiny UFPR-ALPR / Brazil \textsuperscript{\tiny\fontsize{3pt}{6pt}\selectfont{3}}} & {\tiny NA ~/~ NA ~/~ NA ~/ 85.4\% } \\ & & & & \multicolumn{2}{c|}{} & {\tiny OpenALPR / European \textsuperscript{\tiny\fontsize{3pt}{6pt}\selectfont{4}}} & {\tiny NA ~/~ NA ~/~ NA ~/ 85.2\%} 
			
			\\ \cline{2-8} \multirow{-5}{*}{
				\begin{tikzpicture}[remember picture,overlay]
					\node[fill,rectangle,top color={blue!40!white},bottom color={blue!40!white},minimum width=0.64cm,minimum height=4.25cm] (test){\key{\tiny \bf Learning-based}};
				\end{tikzpicture}
			} &
			
			\multirow{2}{*}{\tiny \cite{P15_2020}} & \multirow{2}{*}{\tiny NA} & \multirow{2}{*}{\tiny tiny-YOLOv3 + layout detection} & \multicolumn{2}{c|}{\multirow{2}{*}{\tiny YOLOv3-SPP}} & \multirow{2}{*}{\tiny KarPlate / Korea \textsuperscript{\tiny\fontsize{3pt}{6pt}\selectfont{5}}} & \multirow{2}{*}{\tiny NA ~/~ NA ~/~ NA ~/ 98.2\%} \\ &  & & &  \multicolumn{2}{c|}{} & &
			
			\\ \cline{2-8} \multirow{-26}{*}{
				\begin{tikzpicture}[remember picture,overlay]
					\node[fill,rectangle,top color={white},bottom color={blue!40!white},minimum width=0.64cm,minimum height=3.2cm] (test){};
				\end{tikzpicture}
			}  &
			\multirow{2}{*}{\tiny \cite{P16_2020}} & \multirow{2}{*}{\tiny NA} & \multirow{2}{*}{\tiny tiny-YOLOv3} & \multirow{2}{*}{\tiny NA} & \multirow{2}{*}{\tiny CNN} & \multirow{2}{*}{\tiny NP / Brazil \textsuperscript{\fontsize{3pt}{6pt}\selectfont {3}}} & \multirow{2}{*}{\tiny 99.4\% /~ NA ~/ 98.4\% /~ NA} \\ & & & &  & & &

			\\ \cline{2-8} \cellcolor{black} &  \multicolumn{2}{c}{\multirow{2}{*}{ \footnotesize{\tiny {1-}} \raisebox{-0.5\totalheight}{\includegraphics[width=1.75cm, height=0.45cm]{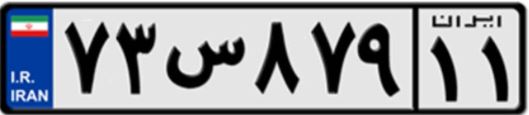}}}} &  \multicolumn{1}{c}{\multirow{2}{*}{ \footnotesize{\tiny {2-}} \raisebox{-0.5\totalheight}{\includegraphics[width=1.75cm, height=0.45cm]{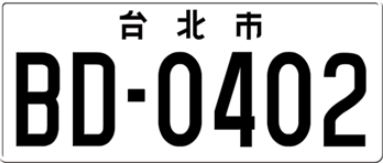}}}} & \multicolumn{1}{c}{\multirow{2}{*}{ \footnotesize{\tiny {3-}} \raisebox{-0.5\totalheight}{\includegraphics[width=1.75cm, height=0.45cm]{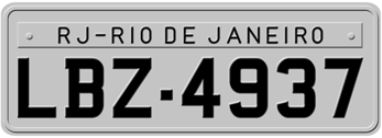}}}} & \multicolumn{2}{c}{\multirow{2}{*}{ \footnotesize{\tiny {4-}} \raisebox{-0.5\totalheight}{\includegraphics[width=1.75cm, height=0.45cm]{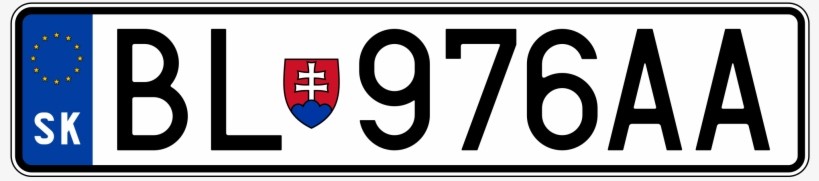}}}} & \multicolumn{1}{c|}{\multirow{2}{*}{ \raisebox{-0.5\totalheight}{\includegraphics[width=5cm, height=0.45cm]{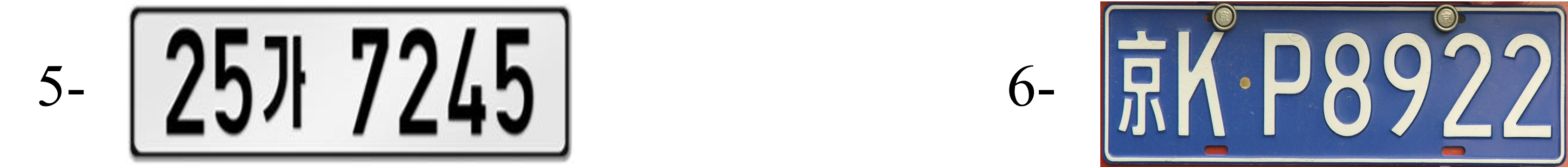}}}} 
			\\ \cellcolor{black} &
			\multicolumn{1}{c}{\multirow{2}{*}{}} &
			\multicolumn{1}{c}{\multirow{2}{*}{}} &
			\multicolumn{1}{c}{\multirow{2}{*}{}} &
			\multicolumn{1}{c}{\multirow{2}{*}{}} &
			\multicolumn{1}{c}{\multirow{2}{*}{}} &
			\multicolumn{1}{c}{\multirow{2}{*}{}} &
			
			\\ \hline

		\end{tabular}
  }	
	
\end{sidewaystable*}

\section{Methodology}
\label{Methodolgy}

As the main components of the proposed method comprise a selective deblur GAN at the preprocessing step and two YOLOv5 structures at the LP and CR steps, we start with a theoretical background on deblurring and object detection.
Following up on the leads mentioned in Sec~\ref{Motivationpart}, blur prevalence and small size of the LPD/CS objects in vehicle imaging systems, we propose the ALPR system of Fig.~\ref{fig:flowchart}. Under the framework presented in Fig.~\ref{fig:ALPR_system}. 

\subsection{Background} 
In this section, we briefly review the background of the components of the proposed method.
\subsubsection{\textbf{Debluring}}
\label{deb}

\par We start with a new definition of the blur effect: Image blur occurs when multiple location instances of a specific object are present in a single image frame. It results from relatively high-speed movement between the camera and the object during the capture period, as in quick-moving objects or shaky cameras. The abrupt pixel intensity changes (i.e., edges) fade into smooth variations as a visual effect. As the objects of interest in ALPR are small LPs and closely-located characters, image blur is one of the most extreme performance degrading phenomena in ALPR, compared to angled view, noisy image, and uneven illumination effects. Therefore, blur addressing at the preprocessing step is specifically needed.
\par Since the blur phenomenon smoothens the edges of the objects of interest in an image, conventionally, differentiation-based methods have been used to mitigate the problem \cite{P18_olddeblurmethod}. More recently, learning-based methods such as hierarchical CNN, Iterative gradient prior-learning and deconvolutional CNN, Pixel-wise motion vector classification, Recurrent NNs, Patch-wise inverse kernel coefficient regression, and NNs with Perceptual loss and Wasserstein distance have been proposed \cite{P19_deblursurvey}. GAN is one of the state-of-the-art deblurring structures. For instance, \cite{P38_superresolution_elsevier} employs two sequential Super Resolution-GAN modules to decrease noise/blur and enhance the resolution of input LP. According to \cite{P20_gan}, GANs create sharp synthesis images that are similar to their real counterparts. The merits of GAN deblurring, to be explained briefly, are our motivation for employing it at the preprocessing step of the proposed ALPR. 
\par One of the efficiency challenges of the latter methods is the unavailability of an adequate dataset. For simulating object motions in synthetic datasets, the uniform blur kernel (to create synthesized images) and the non-uniform blur kernel (by averaging consecutive images from a GoPro camera) have been employed. Unfortunately, the synthetic blur datasets so generated are not sufficiently representative, specifically in a multi-source blur or blur kernel with complex patterns scenarios \cite{P21}. 

\begin{figure}[h!]
	\centering
	\includegraphics[width=\columnwidth]{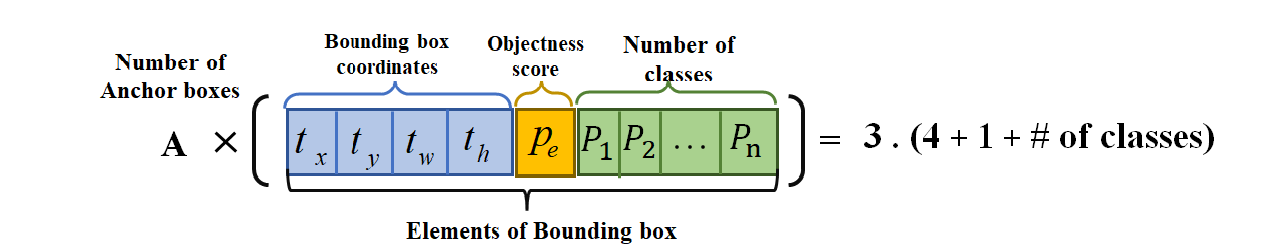}
	\caption{Number of the generated data items per grid cell at each of the three different scales of the
		YOLOv3/v4/v5 outputs.}
	\label{formula}	
\end{figure}

\begin{figure*}[h!]
	\centering
	\captionsetup{justification=centering}
	\includegraphics[width=\textwidth,height=7.5cm]{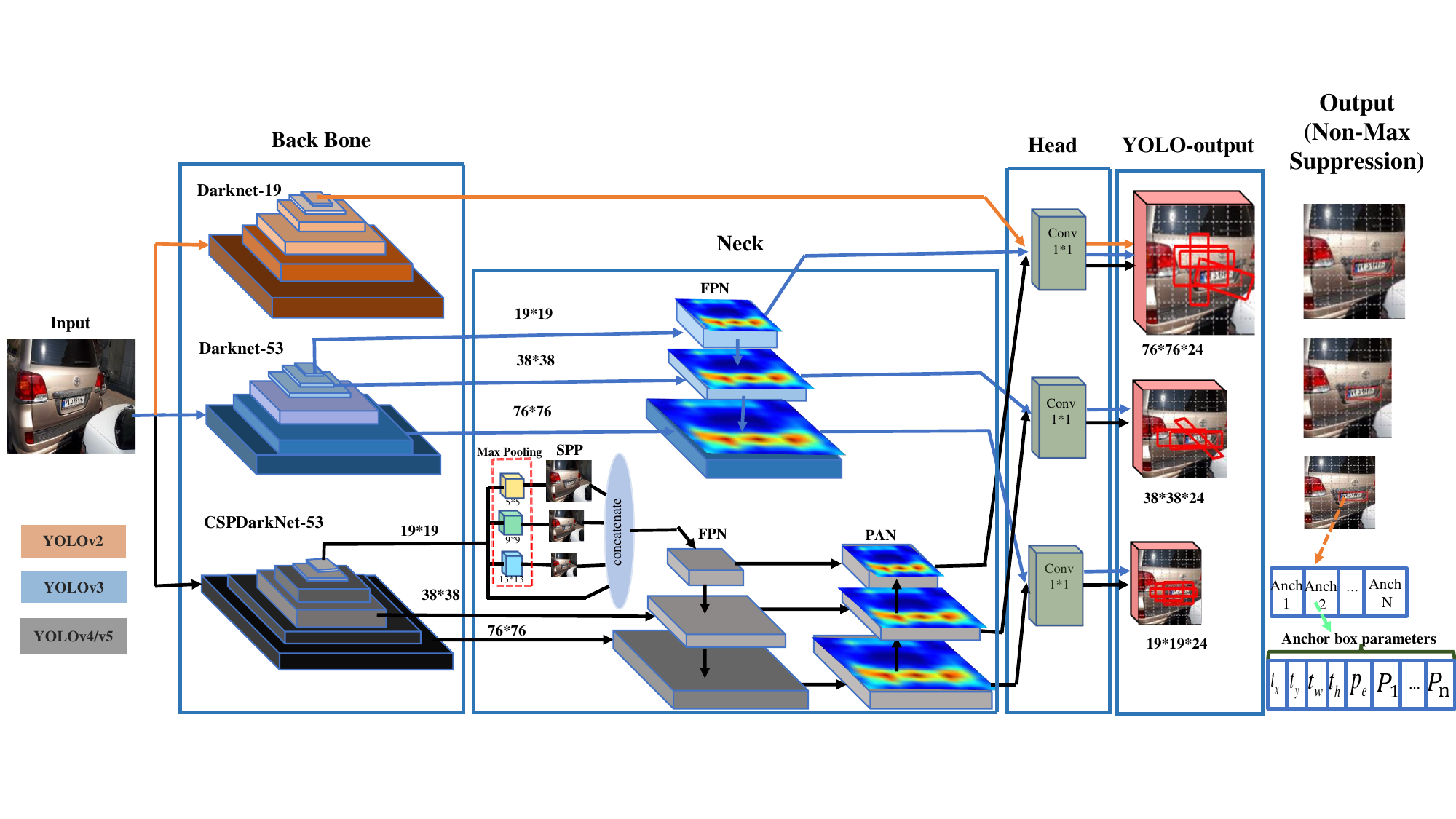}
	\caption{Comparative presentation of different versions of the YOLO structure.}
	\label{YOLO}
	
\end{figure*}

\subsubsection{\textbf{Object Detection}}
\label{obj}
\par In recent years, the advent of CNNs obviates the design of manual feature extractors and inflexible and complicated mathematical operations in earlier machine learning methods. Sliding-window-based detection algorithms are considered among the first proposed CNN methods. A variable-size sliding window moves over an image, and the resulting windowed images are sequentially passed into a CNN for classification. The main disadvantages of these methods are high computational cost and low detection speed. A low-computational-cost sliding-window detection method, Region-based CNN (R-CNN) \cite{P22_RCNN}, applies CNN to a fewer number of windows on selected Regions of Interest (ROI) in the image (called blobs) without any parameter sharing between them, to automatically extract features. Then, an SVM classifier is used to recognize the objects. To overcome the high computational cost of independent processing of blobs, Fast R-CNN \cite{P23_FastRCNN} uses a convolutional feature map of the input image to detect objects. The Faster R-CNN \cite{P24_FasterRCNN} method further improves the speed by replacing fully connected layers with convolutional layers that use parameters in a shared manner. This method uses a CNN instead of segmentation to propose candidate regions.

\begin{table}[h!]
	\scriptsize
	\renewcommand{\arraystretch}{1.4}
	\caption {Comparison of different YOLO versions.} \label{table:YOLO_compare}
	\centering
	\begin{tabular}{c c c c}
		\hline 
		\textbf{Models}  & \textbf{Backbone} & \textbf{Neck} & \textbf{Head}
		
		\\
		\hline
		\tiny \footnotesize{YOLOv2} & DarkNet-19 & & Class and Box subnet

		\\
		\rowcolor[rgb]{0.9,0.9,0.9}
		YOLOv3 & Darknet-53 & FPN & Class and Box subnet

		\\
		
		YOLOv4/v5 & CSPDarknet53 \cite{P32_CSPDarknet53} &  SPP and PAN  & Class and Box subnet
		
		\\
		\hline
	\end{tabular}
	
\end{table}

\par Object detection algorithms conventionally extract objects from an image in two steps, candidate-region (i.e., ROI) specification and classification. The YOLO structure, alternatively, does the object extraction task in only one step achieving a faster recognition \cite{P17_YOLO}. To improve the detection of small objects and draw more accurate bounding boxes around the already detected objects in the input image, YOLOv2 \cite{P25_YOLOv2} is proposed. Besides having a more robust feature extractor, YOLOv2 generates anchor boxes per grid cell. As illustrated in Fig.~\ref{YOLO}, the YOLOv2 structure (and almost equivalently the original YOLO structure) has two main stages, the Backbone stage for feature extraction and the Head stage for prediction. In later versions of YOLO, YOLOv3 \cite{P26_YOLOv3}, YOLOv4 \cite{P27_YOLOv4}, and YOLOv5 \cite{P28_YOLOv5}, as well as employing deeper feature extractor, another intermediary stage, called Neck, is added to process different scales of extracted features from the Backbone. The inclusion of the Neck stage proves helpful in detecting small objects. Moreover, the Neck structure differs in YOLOv3 from YOLOv4 and YOLOv5. In the former, the Neck stage includes only an FPN \cite{P29_FPN} network, which estimates three scales of feature maps to detect small objects better than the older version. In the latter, the lowest feature map scale from the Backbone stage is modified by an SPP \cite{P30_SPP} network before feeding the FPN network. Also, the outputs of FPN are manipulated by a PAN \cite{P31_PAN} network before entering the Head stage. These two changes on YOLOv3 improve the learning capability of YOLOv4 and v5. Table~\ref{table:YOLO_compare} shows the implementation differences between YOLO versions. 

To clarify the outputs of the YOLOv3/v4/v5 structure, a predetermined number of Anchor boxes are considered in each grid cell. The Anchor boxes are bounding boxes with preset sizes that best match the desired objects, and the predicted (i.e., output) bounding boxes are just scaled versions of them whose information is as in Fig.~\ref{YOLO}. So, the number of generated data items per grid cell at each of the three different scales of the YOLOv3/v4/v5 outputs (see the right end part of Fig.~\ref{YOLO}) are calculated according to Fig.~\ref{formula}.

\begin{figure*}[h!]
	\centering
	\captionsetup{justification=centering}
	\includegraphics[width=\textwidth]{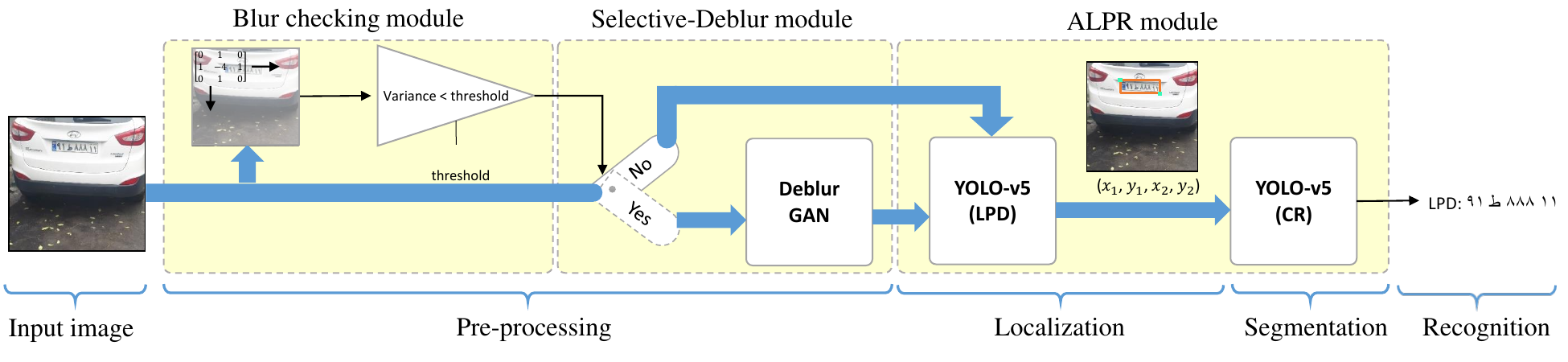}
	\caption{Proposed ALPR system's architecture.}
	\label{fig:flowchart}
\end{figure*}
\subsection{\textbf{The Proposed method}}
This study presents an innovative methodology specifically developed for automatic license plate recognition. The procedural approach is visually illustrated in Fig.~\ref{fig:flowchart}.

\begin{figure*}[h!]
	\centering
	\captionsetup{justification=centering}
	\includegraphics[width=\textwidth]{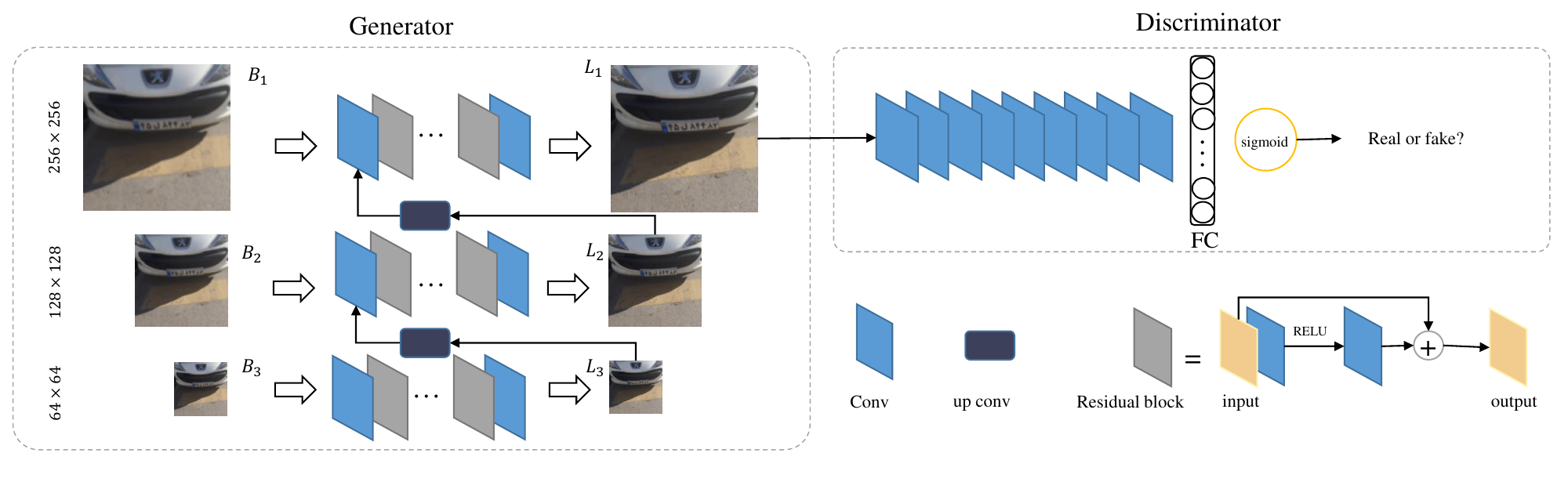}
	\caption{ Deblur-GAN structure.}
	\label{deblur}
\end{figure*}

The method we propose comprises five key stages: input image acquisition, pre-processing, localization, segmentation, and recognition. The image which is captured by the camera is passed to our algorithm as input.
\subsubsection{ pre-processing}
\begin{figure}[h]
	\centering
	\includegraphics[width=\columnwidth]{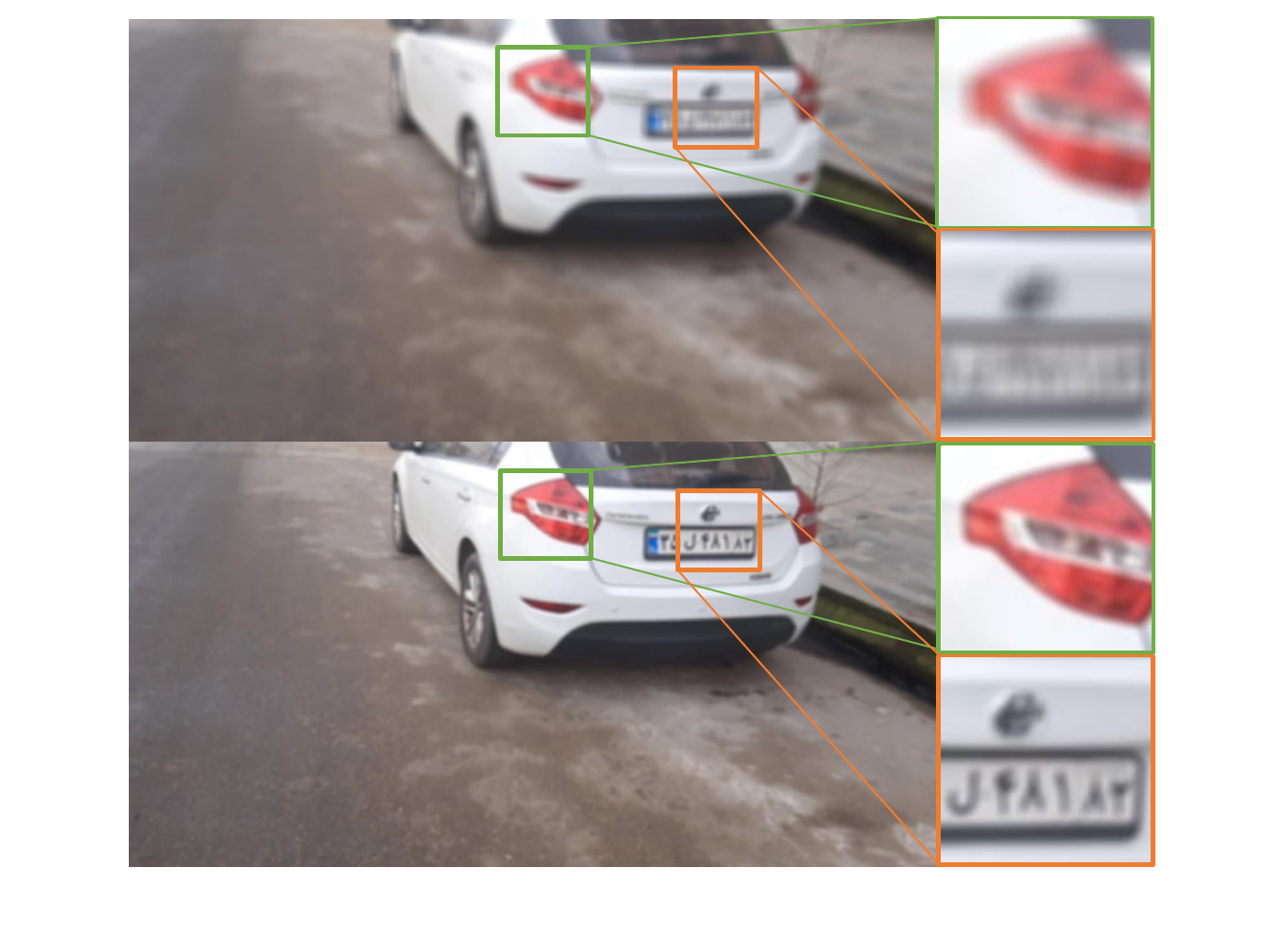}
	\caption{LP detection result without (top) and with (bottom) GAN deblurring.}
	\label{deblur_result}
\end{figure}

in the preprocessing step in Fig.~\ref{fig:flowchart}, we consider a Deblur module.
\begin{itemize}
	\item Blur checking module
	\par Since deblurring of non-blurred images degrades the recognition accuracy, we propose an automatic blur check module to precede and selectively bypass the Deblur module. A Laplacian kernel moves throughout the input image to generate an output image containing the edges. As the blurring phenomenon degrades sharp edges, applying a variance operator on the resultant image will evaluate the input image's edges' intensity. Compared with a carefully selected user-set threshold, the edge intensity indicates whether exposure to blurring has happened. Thus, a higher-than-threshold edge intensity (i.e., a non-blurred image) obviates the need for any deblur treatment and its associated time costs. Alternatively, a lower-than-threshold edge intensity (i.e., blur occurrence) directs the input image to take the Deblurring path (the lower data path in Fig.~\ref{fig:flowchart}) in the proposed ALPR architecture. 
	\item Selective-Deblur module
	\par The deblurring is done using a GAN structure already discussed in Sec.~\ref{deb}. In particular, we use the Deep Multi-scale CNN \cite{P33_deblur-proposed-method} implementation introduced as illustrated in Fig.~\ref{deblur}, wherein the residual blocks in each level of the multi-scale structure provide a deeper structure for extracting features. Also, with a lack of a priori information about the actual resolution of the features, the G is fed with three different resolution scales of the blurry input image, 64$\times$64 (B3), 128$\times$128 (B2), and 256$\times$256 (B1).
	At each level, a lower-resolution deblurred output is then up-convolved and concatenated with its immediate higher-resolution blurry input image to generate a higher-resolution sharp output, leading to the final deblurred output image, S1. Then, the D decides how good the final deblurred outcome is and whether it needs further manipulation of hyperparameters. A real-life input (up) and output (down) of the Deblur-GAN is illustrated in Fig.~\ref{deblur_result}.
\end{itemize}

\subsubsection{Localization}
\par Turning our focus to the ALPR module in Fig.~\ref{fig:flowchart}, there are two YOLOv5 structures in the proposed system; one implements the localization step, and the other one the segmentation step. We choose the YOLOv5 structure because of its ability to deal with small objects and its speed. The YOLOv5 structure was presented in Sec.~\ref{obj} and the only adaptations needed are in the Head stage (see Fig.~\ref{YOLO}). Since a single Conv 1$\times$1 generates each information item, this number also equals the number of required Conv 1$\times$1s at each YOLOv5 output scale. In the proposed ALPR system, we set the number of Anchor boxes to 3. Moreover, the first YOLOv5 is in charge of the LPD task with one object class (i.e., LP).
\subsubsection{segmentation}
On the other hand, the second YOLOv5 structure does the CR with 44 object classes corresponding to 44 characters used in Iranian LPs. As such, the number of required Conv  1$\times$1s in the Head stage of the first and second YOLOv5s are equal to $3.(4+1+1)=18$ and $3.(4+1+44)=147$, respectively.
\subsubsection{recognition}
as we have the coordinates of each character's bounding box, we cropped each character and then concatenated them as a sequential number which is called output CR.

\begin{table*}[]
	\centering
	\caption{Datasets'characteristics.}
	\label{config-dataset}
	\resizebox{\textwidth}{!}{%
		\begin{tabular}{|c|c|c|c|c|c|c|c|c|}
			\hline
			\multicolumn{1}{|l|}{\cellcolor[HTML]{000000}} &
			Capture device &
			Location &
			\multicolumn{1}{l|}{Dataset size} &
			\multicolumn{1}{l|}{Target condition} &
			\multicolumn{1}{l|}{Time} &
			\multicolumn{1}{l|}{Target distance} &
			\multicolumn{1}{l|}{\# of targets/shot} &
			Conditions \\ \hline
			ALPR Dataset &
			\begin{tabular}[c]{@{}c@{}}Sumsung A5\\  Apple iphone5\\  LG X Power\end{tabular} &
			\begin{tabular}[c]{@{}c@{}}Fereydounkenar\\ Tehran\\ Haraz road\end{tabular} &
			6661 image &
			\begin{tabular}[c]{@{}c@{}}Urban routes\\    Highways\\    Parked cars\end{tabular} &
			\begin{tabular}[c]{@{}c@{}}Day\\ Night\end{tabular} &
			1-15m &
			1-6 &
			\begin{tabular}[c]{@{}c@{}}Damages plates\\ Dirty plates\\ Dusty plates\end{tabular} \\ \hline
			Deblur Dataset &
			Samsung A50 &
			Fereydounkenar &
			4051 images &
			Parked cars &
			Day &
			1-10m &
			1-4 &
			\begin{tabular}[c]{@{}c@{}}Blur filter size\\ 7$\times$7,9$\times$9,...,19$\times$19\end{tabular} \\ \hline
		\end{tabular}%
	}
\end{table*}

\begin{table}[!h]
	\renewcommand{\arraystretch}{1}
	\caption {Deblur model's results} \label{table:deblur_result}
	\centering	
	
	\begin{tabular}{c c c c c c c}
		\hline
		\textbf{Blur kernel} & & \textbf{PSNR (dB)} & & \textbf{SSIM} & & \textbf{MSSIM} 
		\\
		\hline  
		
		7$\times$7 && 36.24 && 0.9104 && 0.9786 
		
		\\	\rowcolor[rgb]{0.9,0.9,0.9} 
		
		19$\times$19 && 31.73 && 0.8011 && 0.9214
		
		\\
		
		\hline
	\end{tabular}
	
\end{table}

\section{Datasets}
\label{Dataset}
\begin{figure}[h!]
	\begin{center}
		\subfigure[ Original dataset ]{\label{fig:dataset1a}\includegraphics[width=0.48\columnwidth]{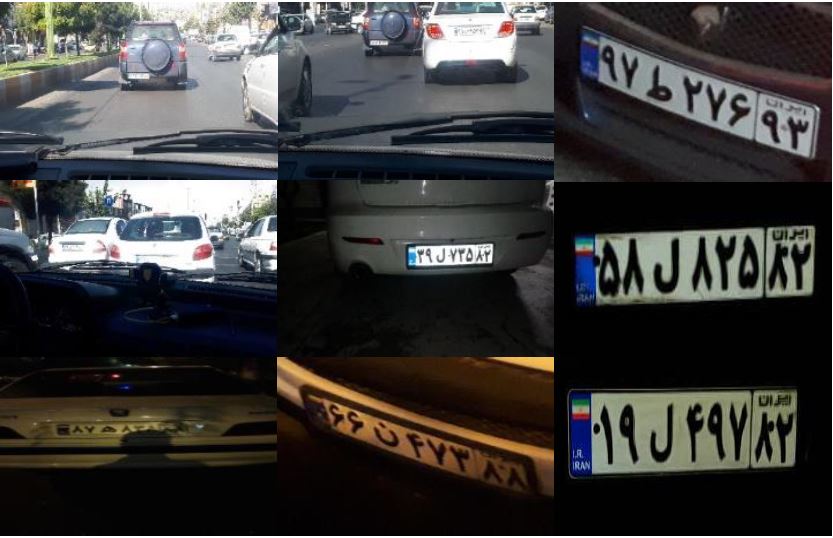}}
		\subfigure[Augmented dataset ]{\label{fig:augmentation}\includegraphics[width=0.48\columnwidth]{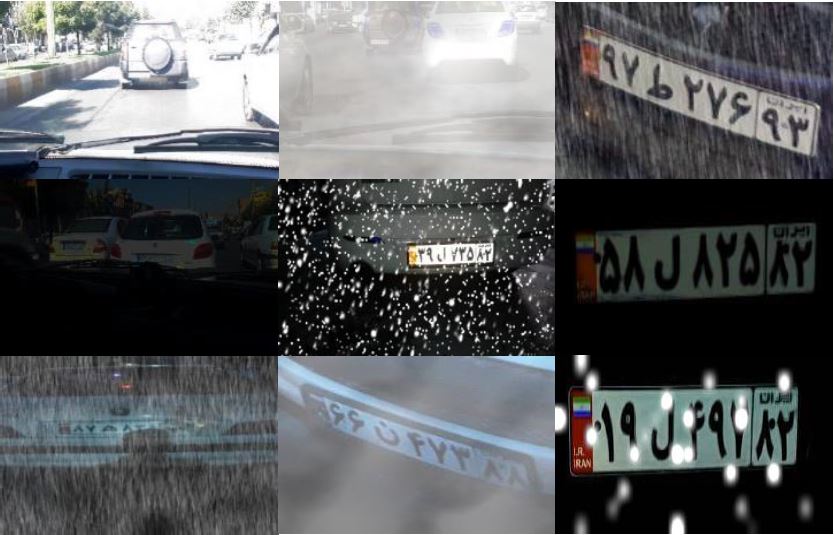}}
		
	\end{center}
	
	\caption{ALPR dataset and its augmentation.}
	\label{ALPR_dataset}
\end{figure}

There are three learning modules in the proposed ALPR architecture in Fig.~\ref{fig:flowchart}, YOLOv5 LPD, YOLOv5 CR, and Deblur GAN. To train these modules, we generated three datasets that exist in \cite{P36_githubme} as follows. For YOLOv5 LPD, we start with the ALPR dataset (see Fig.~\ref{fig:dataset1a} for samples) as described in the first row of Table~\ref{config-dataset}. Then, to simulate different real-life weather conditions, we generated an augmented ALPR dataset using imgaug \cite{P34_imgaug}, a publicly available tool, for rainy, shiny, snowy, foggy, and unlit situations providing 6$\times$6661 images, five per each original image (see Fig.~\ref{fig:augmentation} for samples), totaling 39966 images.

For the YOLOv5 CR dataset, we chose 4808 cropped LP images from the original ALPR dataset (i.e., the first row of Table~\ref{config-dataset}) and then augmented it in the same way as the first dataset, providing 6$\times$4808=28854 LP images. For labeling the two YOLOv5 datasets, makesense tool \cite{P35_makesense} was used. By augmenting the datasets, we aim to improve the features learned by reducing the over-fitting phenomenon and increasing the accuracy of the YOLOv5 models.

\begin{figure}[h]
	\begin{center}
		\subfigure[Sharp images]{\label{fig:sharp}\includegraphics[width=0.45\columnwidth]{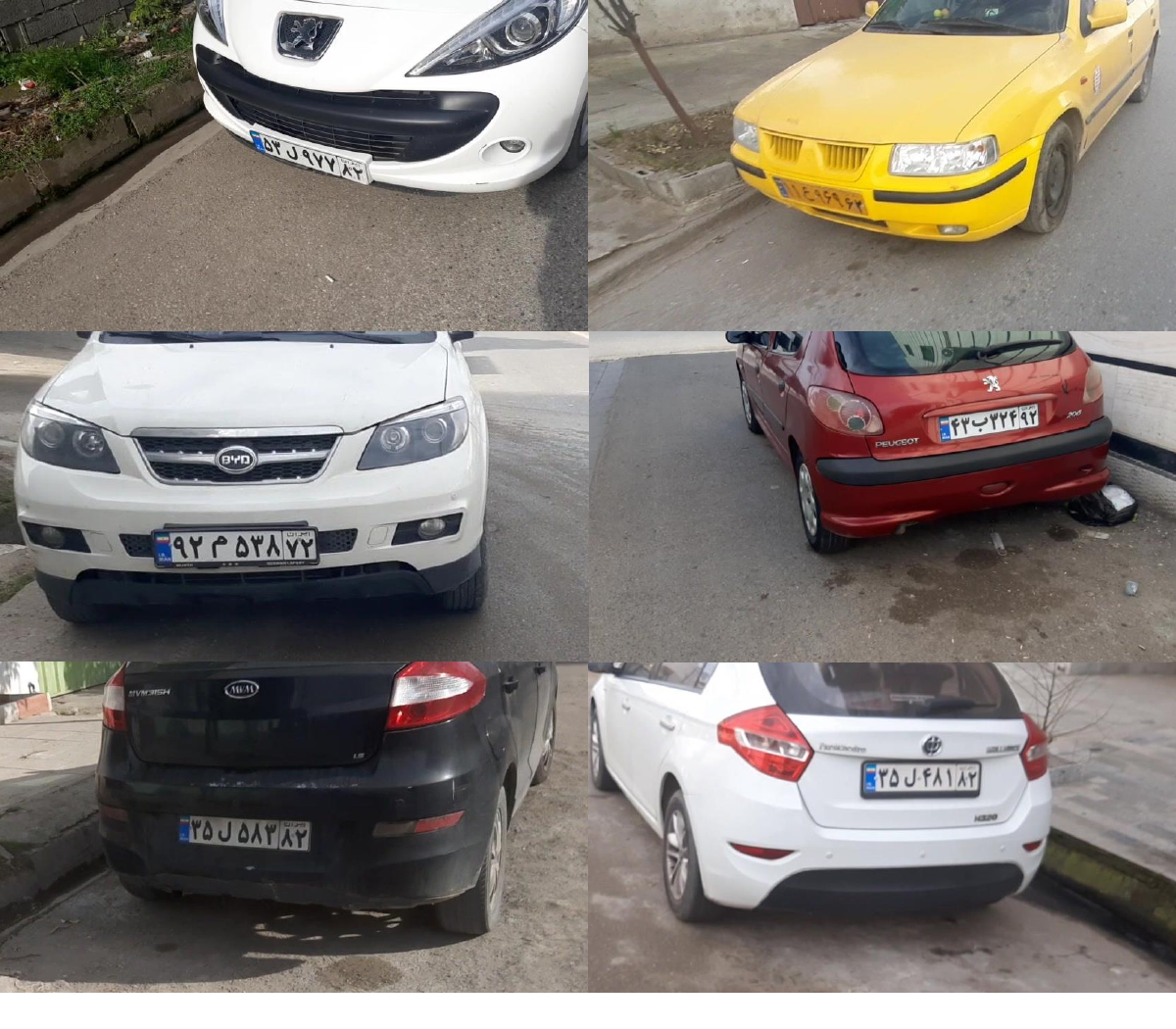}}
		\subfigure[Blurry images]{\label{fig:blur}\includegraphics[width=0.45\columnwidth]{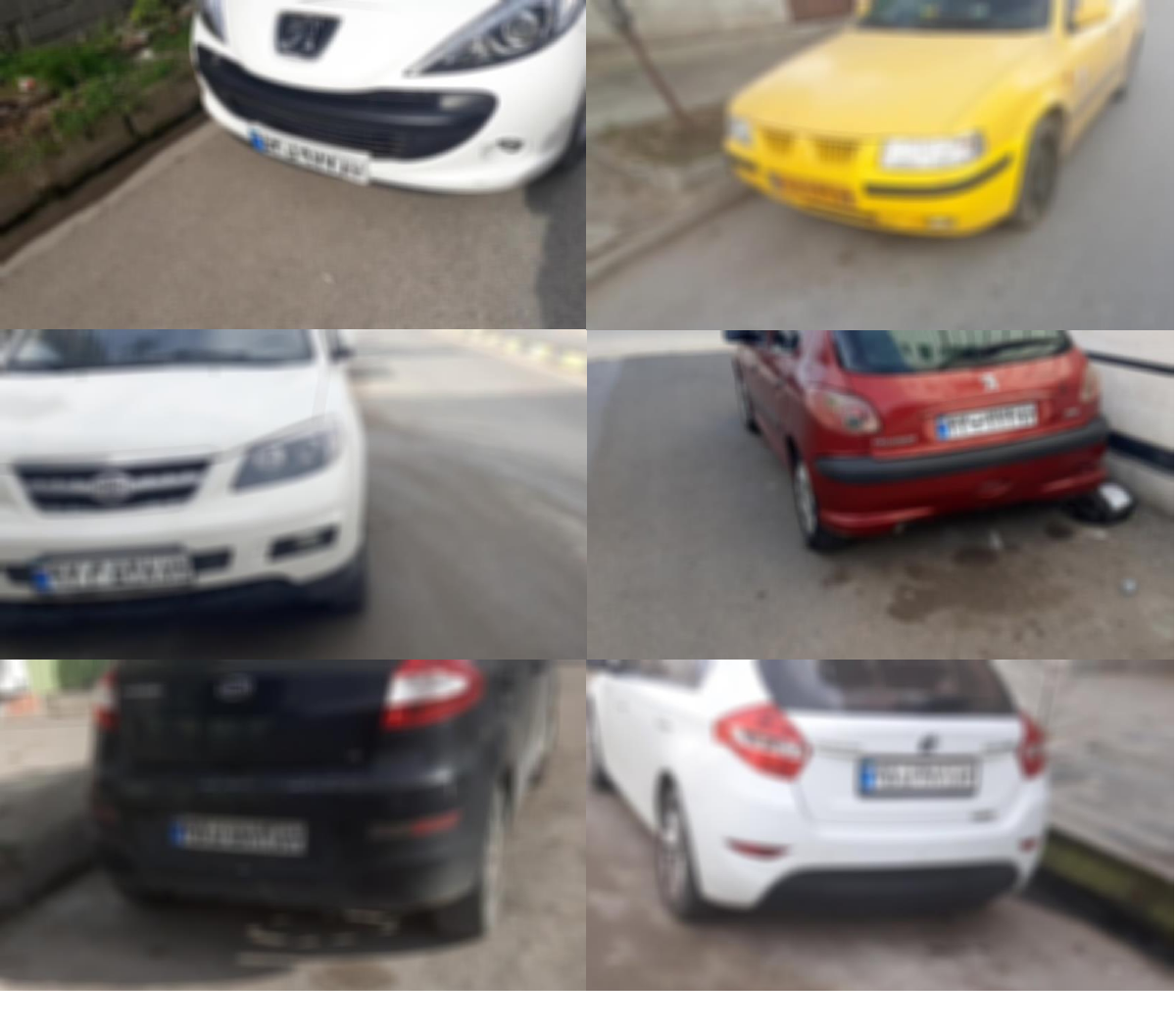}}
	\end{center}
	
	\caption{Deblur dataset.}
	\label{deblur_dataset}
	
\end{figure}

For the Deblur dataset, we generated 78 videos using a moving camera from which 4051 images were captured as per the second row of Table~\ref{config-dataset}. Each image was disturbed by applying (i.e., convolving with) blur filters with different sizes in OpenCV to simulate the camera-shaking phenomenon. Samples from before and after applying the blur filters are shown in Fig.~\ref{fig:sharp} and Fig.~\ref{fig:blur}.

Since applying a larger filter produces a higher disturbance level, we started with a 7$\times$7 filter upwards until it was computationally unaffordable. Then, the largest size achievable was a 19$\times$19 filter with the performance metrics, Peak Signal-to-Noise Ratio (PSNR), Structural Similarity Index Measure (SSIM), and Mean Structural Similarity Index Measure (MSSIM), as per Table~\ref{table:deblur_result}.

\section{Experimental Results}
\label{Simulation}

In this section, we first start with presenting the proposed architecture's end-to-end performance results on YOLOv3 LPD/CR and then, with YOLOv5 LPD/CR modules. Following that, a couple of comparative
remarks with respect to the related state-of-the-art works will be made. The hardware setup and learning parameters of the Deblur-GAN and YOLOv3/v5 models are shown in Table~\ref{tab:models-table}.
\begin{table*}[]
	\centering
	\caption{System's specifications and parameters.}
	\label{tab:models-table}
	\resizebox{\textwidth}{!}{%
		\begin{tabular}{|c|c|c|c|c|c|c|c|c|c|}
			\hline
			\multicolumn{1}{|l|}{\cellcolor[HTML]{000000}}             & Hardware                                                                                               & Optimizer                                                          & \begin{tabular}[c]{@{}c@{}}Learning\\ rate\end{tabular} & \begin{tabular}[c]{@{}c@{}}Learning rate\\  decay\end{tabular} & \begin{tabular}[c]{@{}c@{}}Iou\\ threshold\end{tabular} & \begin{tabular}[c]{@{}c@{}}Confidence\\ threshold\end{tabular} & \begin{tabular}[c]{@{}c@{}}Batch\\ size\end{tabular} & \multicolumn{1}{l|}{Epoch} & \begin{tabular}[c]{@{}c@{}}Train/Validation/Test\\ dataset\end{tabular} \\ \hline
			\begin{tabular}[c]{@{}c@{}}YOLOv3/v5\\ models\end{tabular} & \begin{tabular}[c]{@{}c@{}}GPU : Nvidia Tesla T4\\ RAM : 16 GB GDDR6\end{tabular}                      & \begin{tabular}[c]{@{}c@{}}SGD with \\ momentum 0.937\end{tabular} & 0.01                                                    & 0.2                                                            & 0.6                                                     & 0.3                                                            & 8                                                    & 110                        & 80\%,10\%,10\%                                                             \\ \hline
			\begin{tabular}[c]{@{}c@{}}Deblur-GAN\\ model\end{tabular} & \multicolumn{1}{l|}{\begin{tabular}[c]{@{}l@{}}GPU : Nvidia Tesla T4\\ RAM : 16 GB GDDR6\end{tabular}} & ADAM                                                               & 0.0001                                                  & 0.1                                                            & \_                                                      & \_                                                             & 16                                                   & 400                        & 85\%,5\%,10\%                                                             \\ \hline
		\end{tabular}%
	}
\end{table*}

\begin{table}[!h]
	\scriptsize
	\renewcommand{\arraystretch}{2}
	\caption {LPD and CR Test results} \label{table:YOLO_result}
	\centering
	\begin{tabular}{c c c c c c}
		\hline
		
		\textbf{Model}  & \textbf{\# Images}  & \textbf{Precision} & \textbf{Recall} & \textbf{Time (sec)}
		
		\\
		\hline
		YOLOv3-CR & 1525 & 0.882 & 0.886 & 0.036 
		
		\\ \rowcolor[rgb]{0.9,0.9,0.9}
		YOLOv5-CR & 1525 & 0.976 & 0.986 & 0.026
		
		\\
		YOLOv3-LPD & 3752 & 0.891 & 0.866 & 0.036
		
		\\ \rowcolor[rgb]{0.9,0.9,0.9}
		YOLOv5-LPD & 3752 & 0.95 & 0.954 & 0.026
		
		\\
		\hline
		
	\end{tabular}
\end{table}

\begin{table}[h]
	\centering
	\caption{The proposed ALPR’s performance results.} \label{table:comparison}
	\resizebox{\columnwidth}{!}{%
		\begin{tabular}{c c c c}
			\hline
			{\bf Method} & {\bf \# images} & {\bf Accuracy} & {\bf Time (Sec)} \\
			\hline
			Deblurer (GAN)+YOLOv5 LPD+CR & 300 (75\% blurred) & 87.26\% & 2.052 \\
			YOLOv5 LPD+CR & 300 (75\% blurred) & 45.80\% & 0.052 \\
			\hline
		\end{tabular}
	}
\end{table}

\begin{table}[!h]
	\renewcommand{\arraystretch}{1.4}
	\caption {comparisons (sec)} 
	\label{table:comparison_others}
	\centering
	\begin{tabular}{c c c c c c c}
		\hline
		\textbf{\cite{P13_2020}} & & \textbf{\cite{P8_2017}} & & \textbf{\cite{P11_2020}} & & \textbf{Our method} 
		\\
		\hline  
		0.120 && 0.180 && 0.356 && 0.052 
		\\
		
		\hline
	\end{tabular}
	
\end{table}

\begin{figure*}[h!]	
	\centering
	\includegraphics[width=14.5cm,height=7.19cm]{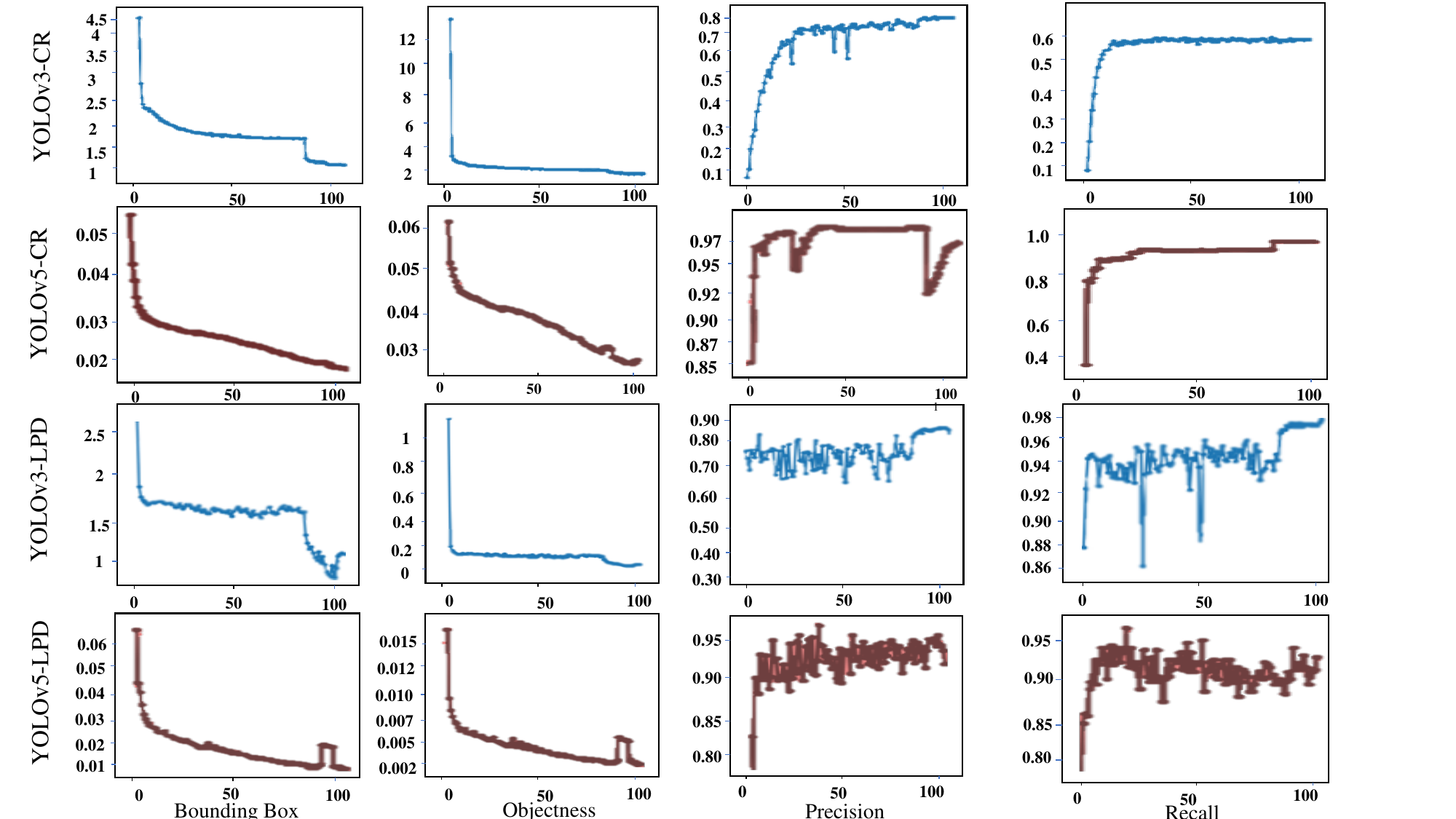}
	\caption{Training convergence behaviour of YOLOv3 and YOLOv5 modules.}
	\label{fig:charts}	
\end{figure*}

\begin{figure}[h!]
	\centering
	\includegraphics[width=\columnwidth]{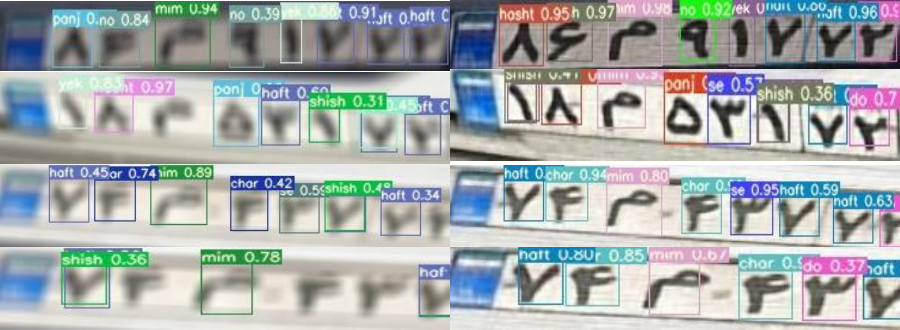}
	\caption{CR results without/with Deblur-GAN (left/right).}
	\label{before-after-deblur}
	
\end{figure}

\subsection{Performance Evaluation}
\label{evaluation}
Focusing on the YOLO-based LPD/CR modules, bounding box and Objectness score (Confidence score) losses are two crucial performance metrics in this application where they represent how accurate the predicted bounding boxes created by the model are and how likely they contain the target objects, respectively.
Furthermore, Precision and Recall, as generic classification metrics, illustrate how well the model is correctly labeling the detected objects and detecting the existing ones, respectively. Figure.~\ref{fig:charts} compares the training convergence behavior of YOLOv5-based and YOLOv3-based LPDs/CRs in terms of the four metrics mentioned above. As is evident, the YOLOv5 modules outperform their YOLOv3 counterparts. 

The test phase results of the employed YOLOv3/v5 LPD/CR are shown in Table~\ref{table:YOLO_result}. The results point to the superiority of YOLOv5-based modules in terms of Precision, Recall, and, very critical to real-time scenarios, Time (i.e., computational complexity). The ranking results of the test phase are in agreement with those of the training phase. The above results support selecting the YOLOv5 structure over YOLOv3 for both LPD and CR modules. Table~\ref{table:comparison} shows the end-to-end performance results of the proposed ALPR architecture, including a selective GAN deblurrer. \\Regarding the impact of employing the deblurrer, the second row shows the performance results over the same dataset when the GAN deblurrer is missing. The significant improvement in the performance points to the critical role of selective deblurring in the preprocessing step. The latter is illustrated in Fig.~\ref{before-after-deblur} wherein the lack of deblurrer leads to a missed, falsely recognized, and/or multiple bounding box per each character of the LP in the CR module. 

\begin{figure}[h!]	
	\centering
	\includegraphics[width=\columnwidth]{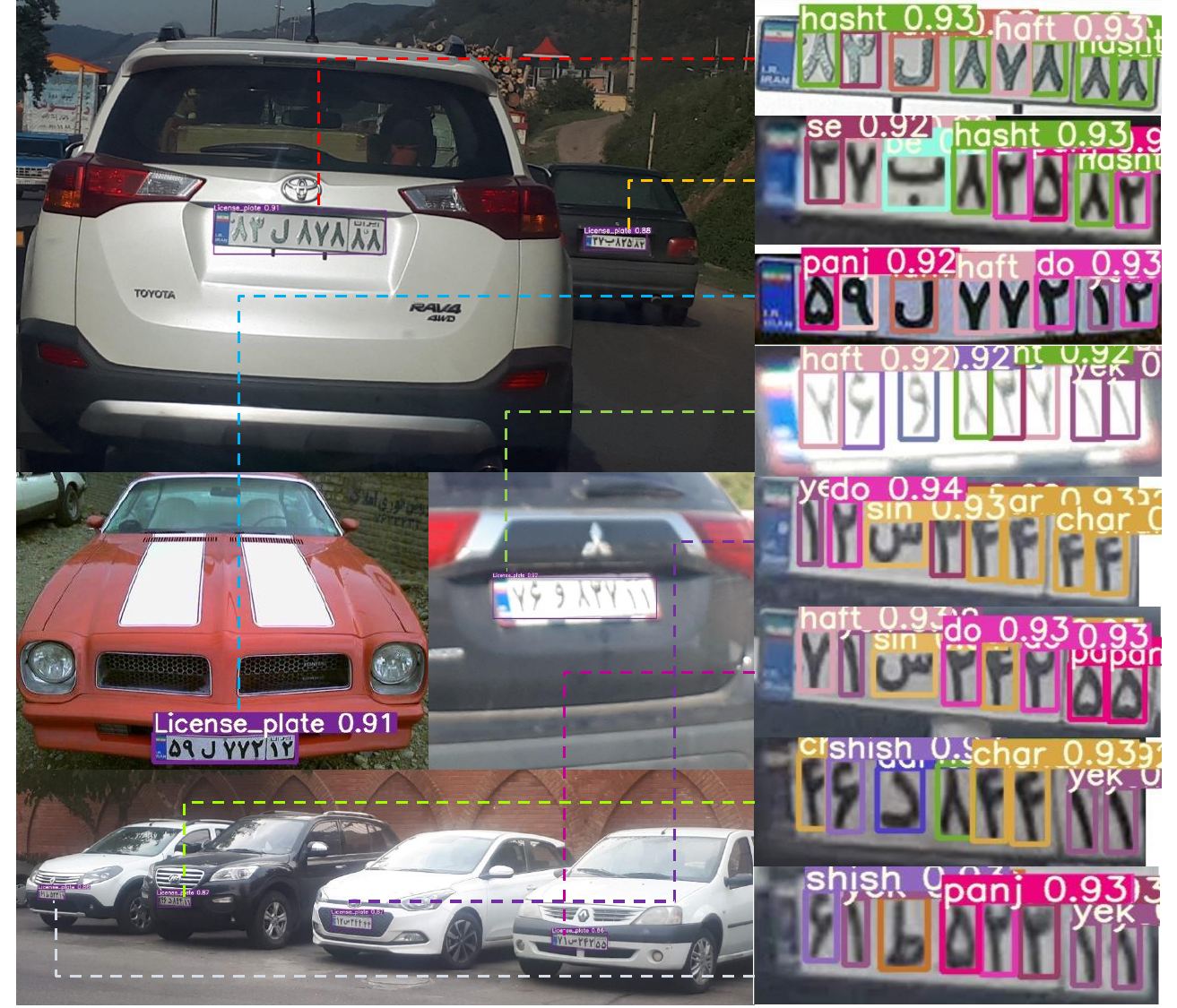}
	\caption{Output results of the proposed ALPR system.}
	\label{fig:ALPR_output}
	
\end{figure}

\begin{figure}[h!]
	\centering
	\includegraphics[width=\columnwidth]{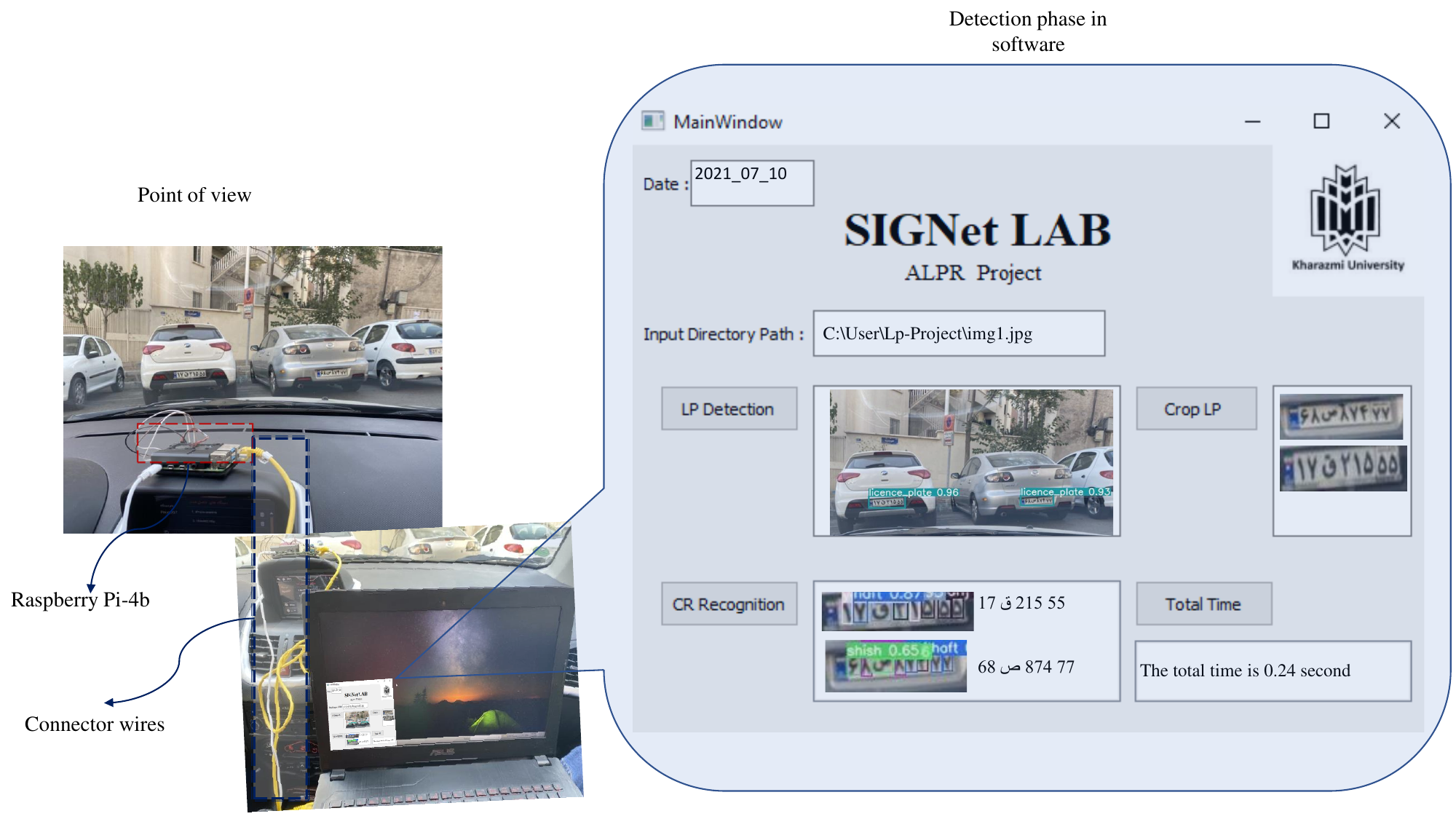}
	\caption{Implementation of the proposed ALPR system using a Raspberrypi-4B hardware.}
	\label{gui}
	
\end{figure}

\subsection{Comparison notes}
\label{comparison note}
In comparison with Iranian ALPR systems, the ALPR system should be trained/tested with diverse complex scenes (image dataset) consisting of multiple LPs under various harsh conditions (distance, angle, partially observable(a portion of LP), blur, noise,...). Based on a dataset in \cite{P37_lastpaper_eccv2018}, nearly covering the whole conditions, is the most comprehensive dataset towards \cite{P39_dataset3,P40_dataset10,P41_dataset14,P42_dataset15,P43_dataset16,P44_dataset17} based on their comparison. However, they collected their images in the parking, which not only have single LPs in the input image but also the not consider the conditions of moving vehicles. Clearly, our work does not have any limitations in the number of LP images in input, whereas, in some works, such as \cite{P37_lastpaper_eccv2018}, the output of the algorithm is designed only to detect seven characters (based on RPnet). Additionally, we consider the real situations causing the blur effects, like handy shakes and moving objects. On the other hand, in \cite{P37_lastpaper_eccv2018}, the RPnet is the best version based on the detection and recognition response time towards other prior object detection algorithms, even YOLO9000 \cite{P25_YOLOv2} until 2018. Nevertheless, from this point, different types of state-of-the-algorithms have been proposed, like different YOLO versions (v4/v5) proposing faster speed in both the recognition and detection phases. In Table~\ref{table:comparison_others} you can see our comparison with respect to other iranian ALPR method, that highlights the least response time between them. Consequently, it motivates us to employ them in our real-time application.  

Figure.~\ref{fig:ALPR_output} gives a visualization of the proposed ALPR system inputs (left) and outputs (right) in different cases of blurred, multiple, and angled-view LPs and also LPs with different visual depths.

To investigate the suitability of the proposal for portable applications, we implemented it using a Raspberry Pi-4B placed on a car's dash, as illustrated in Fig.~\ref{gui}. The output results are shown on a laptop's display connected to the Raspberry Pi-4B. The input images are snapshots generated from the camera in front of the windshield.

\section{CONCLUSION}
\label{Conclusion}

This study examines the role of an ALPR system within ITS, a pivotal component of smart cities. Deep learning algorithms, crucial in addressing the high variability of ALPR systems, play a vital role in system performance. Our investigation focuses on key limitations in ALPR design, including LPD and CR detection times, accuracy in identifying targeted objects, swift reporting of existing LP and CR in input images, and improving defected and blurred images as a pre-processing step. Furthermore, the importance of a well-structured dataset encompassing diverse scenarios such as varying illumination, angles, distances, and weather conditions cannot be overstated.

In summary, this paper demonstrates that two meticulously trained YOLOv5 structures exhibit exceptional accuracy, achieving recognition and detection rates in the high nineties percentile, suitable for real-time and portable applications. Additionally, introducing an early selective GAN-deblurring technique notably enhances overall ALPR accuracy without introducing unnecessary preprocessing drawbacks. To ensure comprehensive training of the YOLO models and the GAN, we generated two extensive datasets: ALPR and Deblur, mirroring close-to-real-life scenarios. Employing data augmentation techniques, including images captured in diverse conditions, ensured well-trained models with reduced overfitting tendencies.

In future work, we aim to explore and incorporate cutting-edge object detection algorithms like YOLO models to enhance efficiency in accurately identifying small objects at considerable distances. Moreover, considering advancements in GAN models, integrating these structures as image-enhancing components could significantly improve input image quality, positively impacting object detection and overall performance.

\section*{ACKNOWLEDGMENT}
We would like to express our gratitude to Navid Pourhadi from Kharazmi University for his help in preparing the datasets and Ehsan Ramezani from Tehran Polytechnic University for hardware implementation assistance.

\section*{REFERENCES}
\def\refname{\vadjust{\vspace*{-1em}}} 

\bibliographystyle{IEEEtran}
\bibliography{ref-sample}{}

\end{document}